# ON NON-MONOTONIC CONDITIONAL REASONING

Hung T. Nguyen
Department of Mathematical Sciences
New Mexico State University
Las Cruces, NM   88003

Abstract. This note is concerned with a formal analysis of the problem of non-monotonic reasoning in intelligent systems, especially when the uncertainty is taken into account in a quantitative way. A firm connection between logic and probability is established by introducing conditioning notions by means of formal structures that do not rely on quantitative measures. The associated conditional logic, compatible with conditional probability evaluations, is non-monotonic relative to additional evidence. Computational aspects of conditional probability logic are mentioned. The importance of this development lies on its role to provide a conceptual basis for various forms of evidence combination and on its significance to unify multi-valued and non-monotonic logics.

Few Keywords. Conditional events, conditional probability, conditional logic, intelligent systems, non-monotonic rules of inference.

## 1. Introduction

Consider the case of automated reasoning under uncertainty in which we wish to take into account the uncertainty in a quantitative way. The Bayesian approach emerges as a popular one, due not only to the fact that probability theory has a firm mathematical foundation, but also to the non-monotonicity of the conditioning operator in probability theory. Generalizations of logic to cope with uncertain information have led to the investigation of the connection between logic and probability (e.g. Hailperin, 1984, also Nilsson, 1986). At a more fundamental level, a logic of implicative propositions, compatible with conditional probability assignments, needs to be established in order to explain conditional reasoning. Such a logic is given in Adams' book (1975). However, in Adams' work, basic objects, namely implicative propositions or conditionals, are taken as primitive in our natural language; syntatic logical operations among conditionals are defined in an ad-hoc manner (simply as one of many possible ways of extending those in Boolean logic); and the basic component for reasoning, namely the logical entailment relation, is defined semantically only for plausible reasoning (i.e. with defaults). Note that, in 1968, Schay published a paper on a mathematical algebra of conditional events! In this paper, we will use our recent works (Goodman and Nguyen, 1988; Goodman, Nguyen and Walker, 1990) to supply a conditioning operator in logic, compatible with that in probability theory, and to show that it is possible to construct mathematically a conditional extension of the first-order logic in which a natural order relation among conditionals is non-monotonic relative to additional evidence. The problem of conditional probabilistic entailment is discussed, and some computational aspects are outlined.

## 2. Non-monotonic rules of inference

Each reasoning procedure in intelligent systems is essentially based upon a logical consequence (or logical entailment) relation in a given logic. In classical first-order logic, the entailment relation $\Rightarrow$ is simply the order relation $\leq$ in the Boolean ring $R$ (of subsets of some set $\Omega$, where $\leq$ is subset inclusion relation, and, by Stone's representation theorem, $R$ represents the collection of propositions in a natural language). To keep things simple, the connectives "and", "or", and "not" are denoted by $\cdot$, $\vee$ and $(\cdot)'$, respectively. Also, for exposition purposes, it is convenient to view R as an algebraic ring with multiplication $\cdot$, addition + (symmetric difference of sets), zero 0 (empty set) and unity 1 ($\Omega$). The partial order on R is: $a, b \in R$, $a \leq b$ if and only if $ab = a$. This $\leq$ in first order logic is obviously monotonic: if $a \leq b$ then $\forall c \in R$, we have $ac \leq b$. More specifically, for $A \subseteq R$, and $Th(A) = \{x \in R : A \Rightarrow x\}$, we say that $\Rightarrow$ is monotonic if $A \subseteq B \subseteq R$ implies $Th(A) \subseteq Th(B)$.

The rule $\Rightarrow$ is non-monotonic if it is not monotonic. By abuse of language, a reasoning procedure is said to be monotonic or non-monotonic according to whether $\Rightarrow$ is monotonic or not. The terms "inference", "reasoning" and "logic", as in probabilistic inference, probabilistic reasoning, and probabilistic logic, should be understood with care! For example, while probability logic is monotonic, the probabilistic reasoning is claimed to capture a non-monotonicity property! It turns out that it all depends on the structure of the syntax of the logic used. To make things precise, by conditional inference (or reasoning), we mean $E \underset{K}{\Rightarrow} a$
(a follows logically from E in the context of K), where $T = <K, E>$ is a theory (Pearl, 1988), K is the knowledge base and E is the evidence. In this framework, $\Rightarrow$ acts like a set-valued function with two arguments (K, E).

The non-monotonicity of $\Rightarrow$ is in general relative to E, i.e. to the addition of facts or evidence. This explains why conditional probability operator is referred to as being non-monotonic: For fixed $a \in R$, the set-function $P(a|\cdot)$ is not monotonic with respect to $\leq$ on R. For non-monotonic reasoning, we refer the reader to McDermott and Doyle (1980), Reiter (1980), McCarthy (1980), Bibel (1986), Geneserth and Nilsson (1987), Pearl (1988).

To bridge the gap between logic and probability, it is necessary to define a conditioning operator in logic, or in Pearl's words "a new interpretation of if-then statements, one that does not destroy the context sensitivity of probabilistic conditionalization" (Pearl, 1988, p. 25). This is consistent with Pearl's words (again) "probabilists should be challenged by the new issues that emerge from the AI experiment" (Pearl, 1988, Preface), or with Grosof "non-monotonic probabilistic reasoning requires us to employ principles for drawing conclusions which properly extend (i.e. go beyond) the axioms of classical probability" (Grosof, 1988). This will be achieved in the next section. Note that this situation is analogous to that of quantum probability (e.g. Gudder, 1988): although quantum mechanics is essentially a probabilistic theory, Kolmogorov's probability model is inappropriate.

## 3. The mathematical conditional extension of logic.

Statements of the form "most b's are a's", "usually, birds fly", "if b then a", etc., are referred to as conditionals or implicative statements, and are symbolized as $(a|b)$ to distinguish them from material implication forms $b \rightarrow a = b' \lor a$.

As in Adams (1975), the basic requirement is that probabilities of conditionals are conditional probabilities. Now, to capture the meaning of conditioning, an object like $(a|b)$ should represent the class of all events $x \in R$ which are "equivalent" to $a$ in the context of $b$, where the "context of $b$" means the class of "consequents of $b$": if $b$ is true, then $x \lor b$ is also true. In other words, the context of $b$ is the filter $R \lor b = \{x \lor b : x \in R\}$. Some algebraic manipulations lead to

$$(a|b) = a + Rb' = \{x = a + rb' : r \in R\},$$

a coset of R.

Let R/R denote the conditional extension of R, i.e.

$$R/R = \{(a|b) = a + Rb', a, b \in R\}.$$

Note that R is contained in R/R via the identification $a \mapsto (a|1)$. It is easy to derive equivalent forms of $(a|b)$. An intuitive one, is an interpretation in "interval analysis": Since $x = a + rb' = ar' \lor ab$, we have: $ab \leq x \leq b' \lor a$, i.e. $(a|b) \subseteq [ab, b \rightarrow a]$. Conversely, if $ab \leq x \leq b' \lor a$, then there exists $r \in R$ such that $x = ab + rb'$, i.e., $x \in (ab|b) = (a|b)$. Thus

$(a|b) = [ab, b \to a]$.

The logical operations on R/R turn out to be

$$(a|b)' = (a'|b)$$

$$(a|b) \cdot (c|d) = (ac|a'b \vee c'd \vee bd)$$

$$(a|b) \vee (c|d) = (a \vee c|ab \vee cd \vee bd) .$$

For inference purposes, the extended order relation to R/R is useful. Define $(a|b) \leq (c|d)$ as $(a|b) \cdot (c|d) = (a|b)$. This happens as soon as $ab \leq cd$ and $c'd \leq a'b$.

It can be shown that if $(a|b) \leq (c|d)$, then, for any probability measure P, we have $P(a|b) \leq P(c|d)$.

Now the conditioning operator $(\cdot | \cdot \cdot)$ is non-monotonic relative to the second argument. Indeed, for a, b, c ∈ R, we have in general $ab \leq abc$ and $a'b \leq a'bc$, so that $(a|b)$ and $(a|bc)$ are not comparable. This fact also explains the non-monotonicity of the conditioning probability operator. See also Dubois and Prade (1988, 1989).

4. **Reasoning with and computational aspects of conditional logic.**

In the reasoning framework $T = <K, E>$, as in Pearl (1988), where $K \subseteq R/R$ and $E \subseteq R$, conditionals of interest are of the form $(a|E)$, for $a \in R$, and in this notation, E stands for the conjunction of all events in E. When K is given together with conditional probabilities $\alpha_i = P(c_i|d_i)$, $i = 1, 2, ..., n$, say, where $(c_i|d_i) \in K$, the value of $P(a|E)$ represents a degree of logical entailment. See also Goodman, Nguyen and Walker (1990).

We will discuss here only a formal computational procedure for $P(a|E)$. It is simply a generalization of Hailperson (1984), see also Nilsson (1986).

Now observe that

$$P(a|b) = P(ab) + P(a|b)P(b') .$$

(*)
Thus, if $\mathcal{P}$ is a finite R-partition of $\Omega$, of size m, say $\mathcal{P} = \{c_1, c_2, ..., c_m\}$, and ab, b' can be expressed in terms of the $c_j$'s, say

$$ab = \sum_{j \in N} c_j , \quad b' = \sum_{j \in M} c_j$$

with $N, M \subseteq \{1, 2, ..., m\}$, then $P(a|b)$ is related to $P(c_j)$, $j \in N \cup M$, in a linear fashion. Hence the problem of computation of, say, $P(a_{n+1}|b_{n+1})$, given $P(a_i|b_i)$, $i = 1, 2, ..., n$, can be described as follows.

From the collection of $2(n + 1)$ events

$$\{a_1, b_1; a_2, b_2; ... ; a_{n+1}, b_{n+1}\} ,$$

form the canonical partition $\mathcal{P}$ of $\Omega$ consisting of all events of the form $a_1^{\alpha_1} b_1^{\beta_1} a_2^{\alpha_2} b_2^{\beta_2} ... a_{n+1}^{\alpha_{n+1}} b_{n+1}^{\beta_{n+1}}$, where $\alpha_j, \beta_j \in \{0, 1\}$, with

$$a_j^1 = a_j , \quad b_j^1 = b_j , \quad a_j^0 = a_j' , \quad b_j^0 = b_j' .$$

All events $a_j b_j$, $b_j'$ are decomposable in terms of the elements of this canonical partition. Let m be size of $\mathscr{P}$, $m \leq 2^{2(n+1)}$, and label these elements as $c_1, c_2, ..., c_m$. In view of (*), a coding or "design" n by m matrix $\Pi$ is defined as $\Pi = [\Pi_{ij}]$, where $\Pi_{ij} = 0, 1$ or $P(a_i|b_i)$ according to $c_j \leq a'_i b_i, \leq a_i b_i$ or $\leq b'_i$. Suppose that $\Lambda_j = P(c_j)$, $j = 1, ..., m$ are solutions of the system of equations:

$$\sum_{j=1}^{m} \Lambda_j \Pi_{ij} = P(a_i|b_i), i = 1, ..., n ;$$

then $P(a_{n+1}|b_{n+1}) = \sum_{j=1}^{m} \Lambda_j \Pi_{n+1,j}$, where as before, $\Pi_{n+1,j} = 0, 1$, $P(a_{n+1}|b_{n+1})$ according to $c_j \leq a'_{n+1} b_{n+1}, \leq a_{n+1} b_{n+1}$ or $\leq b'_{n+1}$.

Of course, as in the case of unconditional probabilities, the solutions for the $\Lambda_j$'s is not unique in general, and bounds for $P(a_{n+1}|b_{n+1})$ can be obtained using linear programming technique. For practical computational purposes, stochastic optimization might be helpful.